\documentclass{article}

    \PassOptionsToPackage{numbers, compress}{natbib}

    \usepackage[preprint]{neurips_2025}

\usepackage[utf8]{inputenc} %
\usepackage[T1]{fontenc}    %
\usepackage{hyperref}       %
\usepackage{url}            %
\usepackage{booktabs}       %
\usepackage{amsfonts}       %
\usepackage{nicefrac}       %
\usepackage{microtype}      %
\usepackage{xcolor}         %
\usepackage{graphicx}
\usepackage{subfigure}
\usepackage{wrapfig}
\usepackage{xspace}

\newcommand{\rparagraph}[1]{\vspace{1.2mm}\noindent\textbf{#1.}}

\definecolor{Gray}{gray}{0.92}
\definecolor{racing-green}{rgb}{0.0, 0.8, 0.6}
\definecolor{awesome-red}{rgb}{1.0, 0.13, 0.32}

\usepackage{booktabs}
\usepackage{url}
\usepackage{makecell}
\usepackage{enumitem}
\usepackage{multirow}
\usepackage{colortbl}
\usepackage{svg}
\usepackage{amsmath}
\usepackage{fancyvrb}
\usepackage{spverbatim}
\usepackage{tcolorbox}
\usepackage{arydshln}
\usepackage{algorithm}
\usepackage{amsmath}

\tcbuselibrary{skins, breakable, theorems}

\title{Scaling and Beyond: Advancing Spatial Reasoning in MLLMs Requires New Recipes}

\author{%
  Huanyu Zhang$^{1,2,3,}$\thanks{Equal Contributions.}~~\thanks{Work done / contribution during internship at Microsoft Research.} 
    ~~Chengzu Li$^{4, * \dagger}$ 
    ~~Wenshan Wu$^{1}$ \\
    \textbf{Shaoguang Mao}
    ~~\textbf{Yifan Zhang}$^{2,3}$ 
    ~~\textbf{Haochen Tian}$^{2,3}$ \\
    ~~\textbf{Ivan Vulić}$^{4}$ 
    ~~\textbf{Zhang Zhang}$^{2,3}$
    ~~\textbf{Liang Wang}$^{2,3}$
    ~~\textbf{Tieniu Tan}$^{2,3,5}$
    ~~\textbf{Furu Wei}$^{1}$\\
    {\href{https://aka.ms/GeneralAI}{https://aka.ms/GeneralAI}}\\
    $^1$Microsoft Research \\
    $^2$Institute of Automation, Chinese Academy of Sciences \\
    $^3$University of Chinese Academy of Sciences\\
    $^4$Language Technology Lab, University of Cambridge\\
    $^5$ Nanjing University
 }

\begin{document}

\maketitle

\begin{abstract}
Multimodal Large Language Models (MLLMs) have demonstrated impressive performance in general vision-language tasks. 
However, recent studies have exposed critical limitations in their spatial reasoning capabilities.
This deficiency in spatial reasoning significantly constrains MLLMs' ability to interact effectively with the physical world, thereby limiting their broader applications.
We argue that spatial reasoning capabilities will not naturally emerge from merely scaling existing architectures and training methodologies. Instead, this challenge demands dedicated attention to fundamental modifications in the current MLLM development approach.
In this position paper, we first establish a comprehensive framework for spatial reasoning within the context of MLLMs. We then elaborate on its pivotal role in real-world applications. Through systematic analysis, we examine how individual components of the current methodology, from training data to reasoning mechanisms, influence spatial reasoning capabilities. This examination reveals critical limitations while simultaneously identifying promising avenues for advancement.
Our work aims to direct the AI research community's attention toward these crucial yet underexplored aspects. By highlighting these challenges and opportunities, we seek to catalyze progress toward achieving human-like spatial reasoning capabilities in MLLMs.
\end{abstract}

\section{Introduction}
\label{sec:intro}

Recent advancement in Large Language Models (LLMs) \cite{achiam2023gpt,dubey2024llama} and Multimodal Large Language Models (MLLMs) \cite{alayrac2022flamingo,chen2024internvl,wang2024qwen2} has heightened discussions regarding artificial systems achieving human-level intelligence \cite{gignac2024defining}. 
While extensive research has evaluated these models' capabilities in specific domains, such as code generation \cite{jimenez2023swe}, mathematical reasoning \cite{mirzadeh2024gsm}, and other tasks \cite{fu2023mme,zhang2024mme}, demonstrating remarkable proficiency in specific areas, a fundamental component of general intelligence, spatial reasoning, remains substantially underdeveloped~\cite{ray2024sat,bubeck2023sparks,yang2024thinking}.

\begin{figure}[t]
    \centering
    \includegraphics[width=\linewidth]{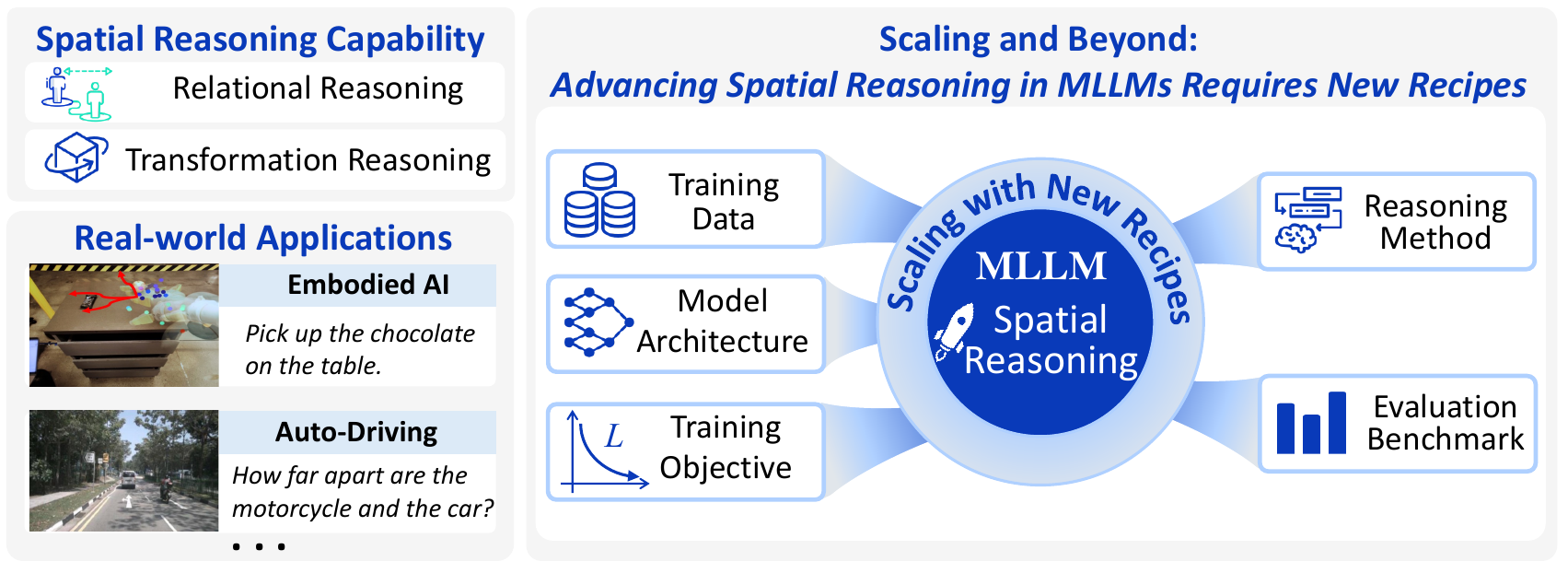}
    
    \caption{By defining spatial reasoning in MLLMs and analyzing limitations in the current recipe, we advocate for new recipes to enhance spatial reasoning, unlocking the potential for applications.}
    
    \label{fig:intro}
\end{figure}

In cognitive science \cite{waller2013handbook} and psychometrics \cite{mcgee1979human}, spatial reasoning (or namely spatial thinking) measures the capability to recall, generate, manipulate, and reason about spatial relations \cite{Gilligan-Lee2022}. 
Within Salthouse's hierarchical structure \cite{salthouse2004localizing}, spatial reasoning emerges as an essential component alongside general reasoning capabilities in determining overall intelligence \cite{Deary2010}. 

However, within the machine learning and deep learning community, systematic investigation of language models' spatial reasoning capabilities remains limited.  
While some studies have examined models' ability to process textual spatial relations \cite{shi2022stepgame,mirzaee2022transfer,li2024advancing}, their scope is constrained compared to the comprehensive framework established in cognitive science.
Recent work with MLLMs has begun exploring this domain more extensively \cite{wangpicture,zhang2024comfort,ray2024sat,li2025imaginereasoningspacemultimodal}. 
Nevertheless, the field lacks a precise definition of spatial reasoning capabilities for these models, impeding rigorous analysis of model behavior and systematic improvement efforts.

To this end, we deliver this position paper to discuss and provide viable pathways in how we can enhance spatial reasoning for MLLMs and hopefully inspire the advancement and traction in this area. 
The emergence of increasingly powerful MLLMs makes this investigation particularly timely, as these models can now process and integrate multimodal information—a prerequisite for human-like spatial reasoning. 
Enhanced spatial reasoning capabilities also contribute to unlocking the potential of MLLMs in downstream tasks and real-world applications that necessitate spatial awareness.
Viewing from a higher level, the impact of advancing spatial reasoning in MLLMs extend far beyond incremental technological improvements. 
It transits from models that passively consume information to systems that can dynamically understand, predict, and interact with physical environments in ways that approach human-like cognitive flexibility.
However, current MLLMs—from open-source to closed-source models—consistently struggle with spatial reasoning tasks that are easy and straightforward for human \cite{ray2024sat,zhang2024comfort,yang2024thinking}. 
This introduces our discussion in this paper: contrary to the assumption that spatial reasoning emerges naturally as a fundamental cognitive skill, we assert that dedicated efforts are necessary to address this challenge.

In this position paper, we argue that \textbf{advancing spatial reasoning in MLLMs requires more than scaling; it requires new design principles and targeted innovations. }
In Section \ref{sec:spatial_reasoning}, we first define two basic MLLM capabilities for spatial reasoning: relational reasoning and transformation reasoning, which extend beyond general visual understanding.
We systematically analyze and identify several key factors that may influence the performance and advancement of MLLMs in spatial reasoning through theoretical analysis and evidence from previous work.  
As illustrated in \figurename~\ref{fig:intro}, the factors include but are not limited to the data and objectives during training (Section \ref{sec:training_data} and \ref{sec:train_obj}), model architectures, reasoning strategies during inference (Section \ref{sec:model} and \ref{sec:reasoning}), etc. 
These components lack the cultivation and design of MLLMs tailored for spatial reasoning, thereby limiting their performance. 
Addutionally, We support our arguments with a detailed case study in Section \ref{sec:case}.

\section{Spatial Reasoning}
\label{sec:spatial_reasoning}

Spatial Reasoning is fundamental to human cognition, encompassing the ability to understand and reason about the spatial relationships among objects, their movements, and interactions with the environment~\cite{chabris2006spatial,Newcombe2024Spatialcog}. 
This sophisticated cognitive skill relies on complex neural mechanisms that integrate visual perception, working memory, executive function, and spatial transformation abilities~\cite{waller2013handbook,knauff2002spatial}. 
While humans naturally develop this capability, achieving this functionality in AI systems presents unique challenges that demand our immediate attention.

In this paper, we focus on spatial reasoning informed by visual cues, a crucial capability associated with visual-spatial intelligence~\cite{yang2024thinking}.
Drawing upon established frameworks from cognitive psychology~\cite{yilmaz2009development} and recent advances in MLLMs~\cite{yang2024thinking}, we propose a comprehensive framework for spatial reasoning in MLLMs.
Specifically, we decompose it into two fundamental capabilities:

\textbf{Relational Reasoning} encompasses the ability to understand and identify spatial relationships between objects, such as relative positions, orientations, and perspective~\cite{chen2024spatialvlm,cheng2024spatialrgpt,zhang2024comfort}. 
This capability captures the static aspect of spatial reasoning, focusing on analyzing spatial configurations.

\textbf{Transformation Reasoning} involves the ability to reason about spatial transformations, including rotations, translations, perspective changes, and other dynamic manipulations of spatial configurations~\cite{ray2024sat,zhang2024comfort,yang2024thinking}. 
This capability addresses the dynamic aspect of spatial reasoning.

By defining these two capabilities, we aim to cover the static and dynamic dimensions of spatial reasoning comprehensively. 
Relational reasoning captures the foundational understanding of ``what is where," while transformation reasoning extends this to ``how things change or could change" within spatial contexts. 
These dimensions closely correspond to the well-established spatial abilities in cognitive psychology: spatial relations and spatial visualization~\cite{yilmaz2009development}.
This alignment not only adapts cognitive theory to the context of MLLMs but also demonstrates its completeness in capturing the essential components of spatial reasoning.

These spatial reasoning capabilities encompass both 2D~\cite{liu2023vsr} and 3D scenes~\cite{hong20233d}, and are fundamental to a wide spectrum of real-world applications where MLLMs interface with spatial information. 
In Embodied AI, Relational Reasoning enables robots to understand spatial relationships between objects, such as recognizing obstacles and determining object positions.
While simultaneously employing Transformation Reasoning, robots can navigate, manipulate objects, and interpret spatial instructions~\cite{hao2024embosr,ishida2024spatial}. 
For Autonomous Driving, spatial reasoning capabilities enable MLLMs to perceive and respond to dynamic environments~\cite{guo2024drivemllm}. 
In Remote Sensing applications, spatial reasoning facilitates the analysis of satellite imagery for urban planning, environmental monitoring, and geographical feature extraction~\cite{zhan2024skyeyegpt,li2024lhrs}.
Within Medical Imaging, both Relational and Transformation Reasoning capabilities are crucial for tasks such as anatomical structure localization and surgical planning from 3D medical scans~\cite{xia2024rule,xia2024cares}. 
In educational applications, MLLMs equipped with spatial reasoning capabilities can assist learners in understanding spatial concepts by leveraging Relational Reasoning for geometric relationships and Transformation Reasoning for dynamic manipulations like rotations~\cite{xing2024survey,ramakrishnan2024does}. 
These diverse applications underscore the critical role of spatial reasoning in enabling MLLMs to understand and interact with the physical world in meaningful ways.

Building upon our framework of spatial reasoning, we conduct a systematic analysis of the current components of MLLM, identify their specific limitations and challenges in spatial reasoning, and propose opportunities for improvement in the following sections. 
Our goal is to catalyze research efforts that will advance MLLMs' spatial reasoning abilities to help real-world applications.

\section{Training}
\label{sec:training}
This section analyzes key factors in MLLM training that impact spatial reasoning: training data, architecture design, and training objective.
We aim to identify opportunities for new recipes and illuminate how targeted innovations can enhance spatial reasoning in next-generation MLLMs.

\subsection{Training Data}
\label{sec:training_data}
In this section, we focus on pre-training and fine-tuning data, as they play a critical role in shaping spatial reasoning capabilities outlined in Section~\ref{sec:spatial_reasoning}.

\rparagraph{Pre-training Data} Pre-training data mainly has two critical functions: aligning different modalities and providing world knowledge~\cite{yin2023survey}.
Recent analyses~\cite{kamath2023whatsup} have revealed limitations in pre-training datasets on spatial relationships, particularly the insufficient spatial annotations. 
Building on \textit{What's UP}, we extend the analysis to additional prominent datasets, including LAION-2B~\cite{schuhmann2022laion}, CC-3M~\cite{sharma2018conceptual}, SBU Captions~\cite{ordonez2011im2text}, and ShareGPT4V-PT~\cite{chen2025sharegpt4v}.
We carefully filtered out spatial prepositions used in non-spatial contexts, such as "under \$25."

\begin{wraptable}[8]{r}{0.5\linewidth}
\renewcommand{\arraystretch}{1.1}
\vspace{-1mm}
    \caption{Frequency of appearance of common spatial prepositions in various pre-trained datasets.}
    \label{tab:pretrain_data}
    \vskip 0.05in
    \centering
    \resizebox{\linewidth}{!}{
    \newcolumntype{C}{>{\centering\arraybackslash}p{0.33\linewidth}} %
    \newcolumntype{D}{>{\centering\arraybackslash}p{0.3\linewidth}} %
    \newcolumntype{E}{>{\centering\arraybackslash}p{0.15\linewidth}} %
    \begin{tabular}{l|cc}
    \hline
      \textbf{Dataset}   & \textbf{\#Samples} & \textbf{Preposition Ratio \%} \\
      \hline
       LAION-2B  & 5.9B & 0.22\% \\
       CC-3M & 3.3M & 0.51\% \\
       SBU Captions & 1M & 1.44\% \\
       ShareGPT4V-PT & 1.2M & 64.71\% \\
    \hline
    \end{tabular}
    }
\end{wraptable}

As shown in Tab~\ref{tab:pretrain_data}, our findings indicate that LAION-2B, CC-3M, and SBU Captions contain notably low proportions of spatial prepositions in their captions. 
This indicates a great gap in their coverage of basic spatial relationships, which significantly constrains the Relational Reasoning capability of MLLMs.
While ShareGPT4V-PT offers more detailed spatial descriptions, its high-quality annotations come at substantial cost.
Furthermore, the limitations become increasingly evident in the context of Transformation Reasoning capability, as current datasets rarely contain descriptions of spatial transformations.
For instance, in \textit{What's UP}~\cite{kamath2023whatsup}, the performance was improved by leveraging specialized spatial relation corpora based on existing datasets, such as COCO~\cite{lin2014microsoft}.
However, fundamental challenges such as perspective transformation remain unresolved.

Beyond basic spatial relation descriptions, other datasets, such as XVLM~\cite{zeng2021xvlm}, have incorporated denser supervision signals like bounding box.
But existing approaches remain constrained by their reliance on natural language for annotating spatial information in images.
Text descriptions primarily capture unary (single object) or binary (two object) relationships~\cite{comșa2023benchmark}, such as ``the cup is on the table" or ``the person stands between two chairs". 
This fundamental limitation makes it inefficient to represent complex spatial configurations involving multiple objects and their intricate spatial relationships. 
When describing sophisticated layouts, natural language descriptions become verbose and potentially ambiguous~\cite{landau2009spatial}.
Looking ahead, we believe the field should explore more sophisticated spatial representation schemes that transcend these limitations. 
Alternative approaches like topological representations could offer more compact and precise ways to encode spatial relationships, potentially revolutionizing how we represent spatial information in MLLMs.

\rparagraph{Fine-tuning Data for Spatial Reasoning} 
Recent work has made significant progress in developing fine-tuning datasets that target different aspects of spatial reasoning capabilities. 
SpatialVLM~\cite{chen2024spatialvlm} performs depth estimation to lift internet-scale images to object-centric 3D point clouds and use it to synthesize VQA data embedded with 3D spatial reasoning supervision.  
SpatialRGPT~\cite{cheng2024spatialrgpt} uses 3D scene graphs to construct template-based spatial descriptions.
These descriptions,  combined with region tags are utilized as input. 
While the above datasets focus on quantitative relationships such as distances and size differences, SAT~\cite{ray2024sat} broadens the scope to dynamic spatial transformations, enhancing both task-specific performance and zero-shot generalization on real-image spatial benchmarks. Complementing these efforts, RoboSpatal~\cite{song2025robospatial} provides egocentric images and 3D scans with spatial annotations to enhance reasoning in robotic settings across diverse reference frames.

Despite these progress, fine-tuning data for spatial reasoning still has a lot of room for improvement.
There is a significant lack of large-scale datasets explicitly designed with spatial reasoning rationales, which hinders the development in complex reasoning.
Therefore, future research should develop more high-quality multimodal reasoning datasets.

\begin{tcolorbox}[top=1pt, bottom=1pt, left=1pt, right=1pt]
    \textbf{Our position:}~\textit{
  Spatial reasoning in MLLMs is hindered by limited spatial annotations in pre-training data, and the absence of explicit spatial reasoning rationales in fine-tuning datasets.
  Therefore, we advocate for (1) improving the quality and coverage of textual spatial annotations and exploring complementary annotation approaches, and (2) incorporating explicit reasoning paths in fine-tuning data to support complex spatial reasoning. 
  }
\end{tcolorbox}

\subsection{Model Architecture}
\label{sec:model}

Having completed analysis of training data, we now examine how model architectures process and reason about spatial information. The flow of spatial information through a MLLM follows a distinct pathway. This pathway begins with initial visual perception through the vision encoder, proceeds through an optional modality alignment connector, and merges in the LLM backbone. Each stage of this processing pipeline presents unique challenges and opportunities for enhancing the model's spatial reasoning capabilities.

\rparagraph{Vision Encoder}
The vision encoder serves as the primary interface for extracting spatial information from visual inputs. 
Current vision encoders, such as Vision Transformers (ViTs)~\cite{Dosovitskiy2020vit}, predominantly adopt transformer-based architectures.
Research in experimental psychology and neuroscience has shown that place and grid cells in the hippocampus and entorhinal cortex contribute to spatial encoding, helping organisms perceive and remember locations~\cite{moser2008place}.
Drawing parallels to artificial neural systems, positional embeddings in vision encoders serve a similar fundamental purpose.
Since the self-attention mechanism in transformer is independent of the token index or positions, the vision encoder depends on position embeddings to provide essential spatial information.

There are two primary methods in position embedding for Vision Transformer: Absolute Positional Embedding (APE)~\cite{Dosovitskiy2020vit,devlin2018bert} and Relative Position Embedding (RPE)~\cite{liu2021swin,shaw2018self}. 
APE encodes the absolute position of each token through sinusoidal functions or learnable embeddings, reminiscent of grid cells.
RPE, similar to place cells, integrates relative position biases into the attention matrix, modeling the relative spatial relationships between tokens.

Nevertheless, both APE and RPE present certain limitations. 
APE lacks flexibility and the ability for extrapolation, whereas RPE restricts the attention module, making it challenging to reflect sufficient positional information to the output~\cite{luo2022your}. 
Therefore, in the visual encoder, APE may struggle to handle spatial transformations such as rotation and translation. 
Similarly, RPE could limit the visual encoder's ability to fully encode and propagate complex spatial information, particularly when dealing with precise distance estimation. 
Swin Transformer~\cite{liu2021swin} explored a simple approach to combine RPE and APE, but results were not promising.
Inspired by the coordinated function of grid and place cells in human cognition, it may be worthwhile to investigate more advanced hybrid positional encoding methods that could better capture and integrate spatial information.

In parallel, recent studies have explored the use of pre-trained 3D encoders~\cite{wang2024omnidrive,shao2024lmdrive,xu2025occllm,yang2025fp3}.
However, these 3D encoders are trained on domain-specific tasks~\cite{liu2022petr, tong2023scene, zhou2023uni3d} and lack the general visual-language understanding ability like CLIP. This mismatch introduces a trade-off between generalization and 3D spatial understanding in MLLMs. A common strategy to mitigate this issue is to adopt a dual-branch encoder design~\cite{ding2024holistic}, which integrates both a general-purpose visual encoder and a 3D encoder. Nonetheless, this approach often leads to feature redundancy and potential representation entanglement, limiting its scalability and efficiency.

\rparagraph{Connector}
The connector, when present, plays a crucial role in preserving spatial information while bridging different modalities. 
It must maintain spatial fidelity while projecting visual features into the language space. 
Specifically, the visual features $f \in \mathbb{R}^{P \times d_v}$ from encoders are processed by the connector $C$, where $P$ is the number of visual patches and $d_v$ is the channel dimension.
The alignment can be formulated as $x = C(f), C: \mathbb{R}^{P \times d_v} \rightarrow \mathbb{R}^{Q \times D}$.
Here, $x\in \mathbb{R}^{Q \times D}$ represents the visual tokens that are input into the LLM, $Q$ is the numbder of visual tokens and $D$ is the hidden size of LLM. 
In a recent study on connector~\cite{lin-etal-2024-preserve},  the authors categorize connectors into two types: feature-preserving connectors, where $P = Q$, and feature-compressing connectors, where $P > Q$.

The feature-preserving connectors can retain detailed information in the visual tokens~\cite{lin-etal-2024-preserve}.
Llava series~\cite{liu2024llava} are one of the popular works in this category. 
While feature-preserving connectors focus on retaining detailed visual information, feature-compressing connectors reduce feature patches for better computational efficiency without losing essential information. 
Various approaches have been proposed to achieve this balance: 
Emu2~\cite{sun2024emu2} employs local average pooling, and BLIP-2~\cite{li2023blip} utilizes the Q-Former, a cross-attention connector with fixed learnable queries for global weighted pooling. 
Besides, HoneyBee~\cite{cha2024honeybee} introduces the C-Abstractor, leveraging convolutional networks for local weighted pooling.
However, the complex connectors like Q-Former may require more aligned data for better results~\cite{lin-etal-2024-preserve}.
Furthermore, the researchers from NVLM~\cite{dai2024nvlm} have found that the cross-attention to latent array in the Perceiver~\cite{jaegle2021perceiver,alayrac2022flamingo} mixes the input image tokens, potentially disrupting the spatial relationships between image patches, which are crucial for spatial reasoning.

These connector designs face a fundamental trade-off between preserving spatial information and computational efficiency. 
Recent work has started to address this challenge - notably, Cambrian~\cite{tong2024cambrian} proposed a novel Spatial Vision Aggregator design that maintains spatial structure during cross-attention operations. 
This innovation demonstrates the potential for architectural improvements in this critical component. 
We advocate for future research to develop novel connector architectures that can better balance this trade-off while explicitly preserving spatial information.

\rparagraph{LLM Backbone}
The language model backbone ultimately bears the responsibility of spatial reasoning using the aligned visual features. 
Its ability to understand and manipulate spatial concepts depends not only on its architecture but also on how effectively spatial information has been preserved through the preceding stages.

The prevalent approach in current MLLMs, which flattens visual tokens into a 1D sequence before applying RoPE-1D~\cite{su2024roformer} alongside text tokens in the LLM backbone, inherently introduces influence to spatial information. 
This linearization of 2D visual information fundamentally compromises the spatial structure of images. 
Recent work like Qwen2-VL~\cite{wang2024qwen2} has made promising progress in addressing this limitation through M-RoPE, which decomposes positional encoding into temporal, height, and width components. 
The superior performance brought by M-RoPE on various multimodal tasks suggests that proper position embedding design can significantly impact the model's ability, including spatial reasoning.
Therefore, future work should explore better position embeddings along this path to enhance spatial reasoning in MLLMs.

In the backbone, various attention masks can be utilized to control the information flow from different modalities.
A typical one is the causal mask, which requires each token to only depend on prior context for generation.
However, this approach may constrain the model's ability to process spatial relationships, as visual spatial information inherently requires bidirectional attention patterns.
Recent works~\cite{beyer2024paligemma,steiner2024paligemma2} have demonstrated improved performance on spatial reasoning benchmark VSR~\cite{liu2023vsr} by allowing more tokens to participate in the reasoning process.
This suggests that the application of causal masking may be suboptimal. 
Therefore, we encourage future studies to explore attention masking strategies that better align with the non-sequential nature of visual spatial information.

\begin{tcolorbox}[top=1pt, bottom=1pt, left=1pt, right=1pt]
    \textbf{Our position:}~\textit{
The spatial reasoning capabilities of MLLMs are constrained by several critical components within their current architectures, from the vision encoder to the LLM backbone.
  To push the current boundaries of spatial reasoning in MLLMs, we call for rethinking these key architectural elements and developing new designs.
  }
\end{tcolorbox}

\subsection{Training Objective}
\label{sec:train_obj}

In this section, we explore the training dynamics of MLLMs, emphasizing their limitations in addressing spatial reasoning. 
We first introduce the loss function design and outline the training modalities. 
Then, we highlight the shortcomings of training paradigms in improving spatial reasoning.

\rparagraph{MLLMs with Discrete Visual Tokens}
MLLMs with discrete visual tokens use visual tokenizer to project visual features into discrete token representations. 
Visual tokens are then fed into language model backbone along with language tokens for reasoning. 
However, this integration to Transformer backbone primarily focuses on semantic reasoning, which is optimized by cross-entropy loss, without explicitly addressing the nuances of spatial reasoning. 
This approach presents several challenges as follows. 
\emph{1) Mismatch in Token Distributions}: Visual tokens are treated identically to text tokens. 
Unlike text tokens, which follow Zipf's law \cite{Piantadosi2014ZipfsWF}, visual tokens originate from an entirely different distribution and thereby are not yet to be determined to be inherently suitable for unified modeling with language tokens in the same way \cite{chan2024analyzing}.
\emph{2) Absence of Spatial Alignment}: 
Cross-entropy loss primarily aligns semantic content between text and visual tokens but fails to incorporate explicit signals for spatial layout understanding. 
Recent works, such as XVLM \cite{zeng2021xvlm}, address this limitation by introducing localized captions and bounding boxes to facilitate fine-grained spatial-semantic alignment. 
However, these methods heavily rely on the granularity of annotated datasets, which may be influenced in cases where annotations are sparse or noisy.
\citet{premsri2024neuro} have explored neuro-symbolic training methods that embed symbolic rules into loss design, aiming to mitigate the lack of spatial constraints. 
However, it remains constrained to the textual domain, lacking the multimodal context required for richer spatial understanding.

\rparagraph{MLLMs with Continuous Denoising Process}
Continuous modeling, as seen in diffusion denoising approaches, introduces another novel paradigm of MLLMs for handling spatial reasoning. 
Notable advancements in this area include Transfusion \cite{zhou2024transfusion} and LatentLM\cite{sun2024multimodal}, both of which combine discrete token modeling for text with a continuous diffusion process for generating images. 
While diffusion loss guides the denoising process, its effectiveness in capturing spatial layouts remains unclear without spatial constraints (e.g. bounding boxes) \cite{patel2024enhancing}. 
This hybrid approach highlights the potential of leveraging both discrete and continuous representations, but further studies are required to determine how well image-level representations retain detailed spatial relationship.
Exploring spatially aware loss functions or additional supervisory signals could provide significant insights.

\rparagraph{Training Modality}
Another key aspect for training MLLMs lies in the modality where the loss is calculated.  
Most MLLMs calculate loss solely based on textual tokens, limiting the flow of information from other modalities. 
In contrast, recent advancements in multimodal generation incorporate image data into the optimization process, enabling richer cross-modal interactions. 
Focusing on dynamic spatial reasoning, \citet{li2025imaginereasoningspacemultimodal} compare a text-only setup with an interleaved text-image scenario where images do not contribute to the loss. 
Interestingly, even when the model is optimized solely on text tokens, interleaved text-image training outperforms or matches text-only training.
When supervision signals are derived from both text and visual tokens, the model demonstrates superior performance, generating more effective visual and verbal reasoning traces.
This highlights the importance of leveraging multimodal information to enhance supervision signals, in addition to careful loss function design.

\begin{tcolorbox}[top=1pt, bottom=1pt, left=1pt, right=1pt]
  \textbf{Our position:}~\textit{
  Current training objectives in MLLMs face key limitations: cross-entropy loss fails to capture spatial constraints, while multimodal approaches struggle with distribution mismatch between text and visual tokens and lack explicit spatial supervision. 
  }
\end{tcolorbox}

\vspace{-2mm}
\section{Evaluation Benchmark}
\label{sec:eval}
\vspace{-2mm}
Robust evaluation benchmarks are essential for advancing spatial reasoning capabilities in MLLMs, as they not only measure current abilities but also highlight critical gaps and guide future research directions. 
Several benchmarks~\cite{mirzaee2021spartqa,shi2022stepgame,rizvi2024sparc,li2024advancing} focused on text-only scenarios, where models were evaluated purely through language descriptions. 
While these benchmarks provided initial insights, the emergence of MLLMs has necessitated more comprehensive evaluation frameworks that can assess spatial reasoning across different modalities and scenarios.

In this section, we review existing benchmark studies on spatial reasoning in MLLMs. 
Based on the presence of changes in spatial configuration, we divide the tasks in existing benchmarks into the following two categories:
\textit{Static Spatial Reasoning} and \textit{Dynamic Spatial Reasoning}. 
The former focuses on understanding fixed spatial relationships and configurations, while the latter involves reasoning about dynamic scenarios, such as movements and transformations. 
\vspace{-2mm}
\subsection{Static Spatial Reasoning}

VSR~\cite{liu2023vsr} pioneered comprehensive evaluation across 66 spatial relation types, revealing significant gaps between early VLMs and human performance, and highlighting the crucial role of positional encodings. 
What'sUp~\cite{kamath2023whatsup} in contrast, narrowed its scope to household object pairs in spatial relationships. 
It uncovered models' preference for color attributes over spatial relationships, highlighting specific limitations in spatial understanding.
Similarly, CV-Bench~\cite{tong2024cambrian} evaluated the fundamental 2D (\textit{Spatial Relationship} and \textit{Object count}) and 3D (\textit{Depth Order} and \textit{Relative Distance}) visual understanding tasks of MLLMs.
Beyond 2D tasks, SpatialVLM~\cite{chen2024spatialvlm} introduced 3D spatial VQA tasks incorporating both qualitative and quantitative assessments, while SpatialRGBT-Bench~\cite{cheng2024spatialrgpt} further advanced the evaluation landscape by providing ground-truth 3D annotations across diverse environments.
Collectively, these benchmarks have demonstrated that using data with corresponding spatial annotations for fine-tuning can improve the model's performance on associated tasks.

In addition to general benchmarks, other works have focused on more specialized tasks, yielding deeper insights.
For instance, \citeauthor{wangpicture} introduced three novel VQA benchmarks (\textit{Spatial-Map}, \textit{Maze-Nav}, and \textit{Spatial-Grid}), revealing following findings:
extra visual input offers limited improvement over MLLMs' LLM backbone,
and MLLMs tend to rely less on visual information when sufficient textual context is available.
TopViewRS~\cite{li2024topviewrs} specifically examined spatial reasoning from top-view perspectives, revealing a substantial performance gap between MLLMs and humans.
Additionally, COMFORT~\cite{zhang2024comfort} systematically assessed Frames of Reference (FoR) using 3D images.
Their findings showed that current MLLMs struggle with robustness and consistency in FoR tasks.
Specifically, most models exhibit a preference for reflected coordinate transformations and egocentric relative FoR but face challenges in adapting flexibly to other FoRs.
3DSRBench~\cite{ma20243dsrbench} assesses MLLMs’ 3D spatial reasoning capabilities and reports that even state-of-the-art models perform poorly and scaling laws for MLLMs are not effective for 3D spatial reasoning.

\vspace{-2mm}
\subsection{Dynamic Spatial Reasoning}

The recent work SAT~\cite{ray2024sat} goes beyond static questions to more dynamic tasks, including \textit{Object Movement}, \textit{Goal Aiming}, \textit{Action Consequence}, \textit{Egocentric Movement} and \textit{Allocentric Perspective}.
The results demonstrated that stronger closed-source models (GPT4-o~\cite{achiam2023gpt} and Gemini-1.5-pro~\cite{team2023gemini}) and spatially-tuned models (Robopoint~\cite{yuan2024robopoint} and Cambrian-1~\cite{tong2024cambrian}) still struggle on dynamic tasks despite performing well on static.
In particular, the perspective tasks involving Transformation Reasoning capability remain challenging even for strong models like GPT4-o.
iVISPAR~\cite{mayer2025ivispar} introduces an interactive visual-spatial reasoning benchmark designed to systematically evaluate MLLMs as agents in dynamic environments. Experimentation results demonstrate that while state-of-the-art VLMs can handle basic spatial reasoning tasks, they face significant difficulties with more complex scenarios, especially in 3D environments.

Another recent study VSI-Bench~\cite{yang2024thinking} evaluated dynamic spatial reasoning tasks through a video-based benchmark.
One of their key observations is that the best-performing MLLM use global spatial context and reasoning to infer correctly, indicating MLLMs may develop an implicit world model.
And the error analysis reveals that almost 50\% of errors stems from egocentric-allocentric transformation, indicating a lack of Transformation Reasoning capability.
Additionally, linguistic prompting techniques like CoT~\cite{wei2022chain}, although effective in language reasoning and general visual tasks, are harmful for spatial reasoning tasks.
They also leveraged cognitive maps~\cite{tolman1948cognitive} to explore how MLLMs remember spaces, finding it forms a series of local world models from the video, rather than a unified global model. 
In addition, SPACE~\cite{ramakrishnan2024does} and VisFactor~\cite{huang2025visfactor} evaluate whether MLLMs demonstrate spatial cognition and find that even frontier models perform near chance level.

Despite the comprehensive evaluation of MLLMs' relational and transformation reasoning abilities across various benchmarks, there remains a gap in correlating these abilities with real-world applications, such as Embodied AI. 
While current benchmarks provide valuable insights, they lack empirical and theoretical investigations that demonstrate how better spatial reasoning improves application performance. 
We believe that future research should bridge this gap, providing a deeper understanding of how stronger spatial reasoning in MLLMs can lead to better outcomes in practical applications.

\begin{tcolorbox}[top=1pt, bottom=1pt, left=1pt, right=1pt]
  \textbf{Our position:}~\textit{
  Existing benchmarks lack a direct correlation with real-world applications like Embodied AI.
  We advocate for an evaluation framework that provides investigations to establish a connection between enhanced spatial reasoning in MLLMs and improved application performance.
  }
\end{tcolorbox}

\vspace{-2mm}
\section{Reasoning Method}
\label{sec:reasoning}
\vspace{-2mm}
Current research can be categorized into three groups based on whether reasoning traces are generated and whether external tools are introduced during reasoning.

Direct reasoning involves the model directly generating the answer.
Instead of directly outputting the answer, recent studies have shown that scaling test-time compute \cite{snell2024scalingllmtesttimecompute,li202512surveyreasoning} improves reasoning performance. 
This is achieved through methods such as search algorithms or strategies like CoT reasoning \cite{wei2022chain, zhang2023multicot}.
Within spatial reasoning, the effectiveness of CoT reasoning has also been extensively explored. 
For example, \citet{li2024topviewrs} demonstrated its utility in eliciting spatial reasoning. 
Besides, \citeauthor{wu2024minds} enhanced performance in grid-based dynamic spatial reasoning tasks by representing reasoning traces in an ASCII-style format.
SpatialCoT~\cite{liu2025spatialcot} improves action generation by introducing a two-stage framework that first aligns inputs with spatial coordinates and then performs chain-of-thought spatial grounding.
However, in multimodal spatial reasoning, textual descriptions can be verbose and sometimes imprecise in describing the spatial relationships in the image. 
MVoT~\cite{li2025imaginereasoningspacemultimodal} introduces a novel reasoning paradigm which extends reasoning traces from textual to multimodal representations. 
This opens new direction and opportunities for multimodal native models to tackle the problem of spatial reasoning, with more straightforward illustrations during reasoning. 
Reinforcement learning has been applied across a wide range of vision-related tasks~\cite{shen2025vlm,huang2025vision}, especially given the rise of GRPO as in DeepSeek-r1 \cite{guo2025deepseek}. 
Visual Planning \citep{xu2025visual} takes a step forward in using RL to optimize visual-centric planning trajectory, achieving better performance than language-reasoning baselines. 
Nonetheless, the role of RL in improving spatial reasoning remains underexplored, highlighting a valuable avenue for future investigation.

In addition to utilizing the reasoning capability of a single model, recent advancement in multimodal agentic reasoning unlocks the potential of incorporating multiple models, each acting as an expert in specific tasks, to improve spatial reasoning performance. 
For example, Image-of-Thought prompting \cite{zhou2024imageofthoughtpromptingvisualreasoning} introduces a spatial ruler to accurately perceive spatial relationships and address spatial positioning queries. 
Similarly, SpatialBot~\cite{cai2024spatialbotprecisespatialunderstanding} enable vision-language models (VLMs) to call an external API for depth estimation, enhancing the understanding of spatial information in input images.
MM-ReAct \cite{yang2023mmreactpromptingchatgptmultimodal} employs vision expert models within a ReAct-style pipeline to gather observations that support decision-making.
Despite these efforts, research on designing agentic workflows specifically for spatial reasoning remains limited. 
Future directions could explore the agentic workflow tailored for spatial reasoning, for example, either by structuring the spatial relations with scene graph \cite{zhong2021learning}, or by imagining the spatial transformation with MVoT-alike systems. 

\begin{tcolorbox}[top=1pt, bottom=1pt, left=1pt, right=1pt]
  \textbf{Our position:}~\textit{
  While traditional spatial reasoning has relied on direct prediction and text-based traces, recent exploration of multimodal reasoning traces and specialized reasoning agents shows promise but remains underexplored.
We advocate for two promising directions: (1) expanding the use of multimodal reasoning traces, and (2) developing specialized multi-agent systems with diverse capabilities for different spatial reasoning tasks.
  }
\end{tcolorbox}

\section{Case Study}
\label{sec:case}
\vspace{-2mm}
\begin{figure*}
    \centering
    \includegraphics[width=\linewidth]{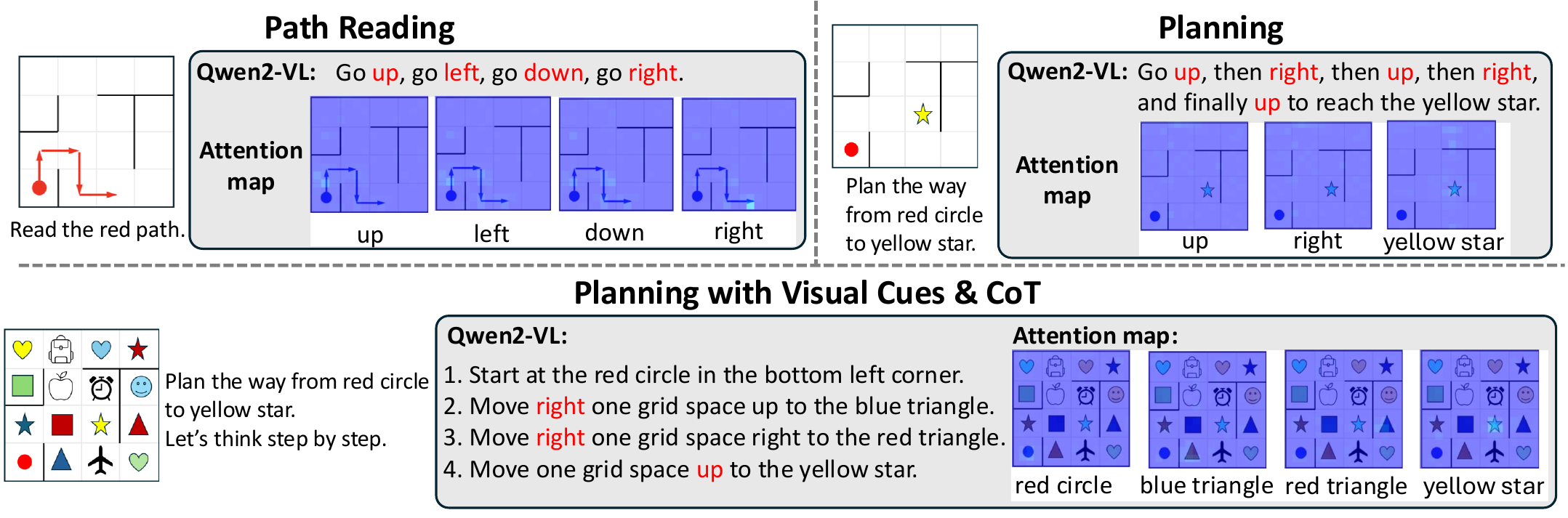}
    \vspace{-5mm}
    \caption{The responses and attention visualization of Qwen-2VL on path reading and planning tasks. 
    }
    \label{fig:case_study}
    \vspace{-3mm}
\end{figure*}

To further illustrate the limitations of current MLLM, we designed two representative tasks, as illustrated in the top section of \figurename~\ref{fig:case_study}. 
In the \textit{Path Reading} task, the model is provided with an image of a maze and asked to describe the path in it. 
The \textit{Planning} task goes a step further by requiring the model to plan a path from a given start point to a specified goal in the maze. 
The detailed prompts can be found in Appendix~\ref{app:case}.
In this case study, we employ Qwen2-VL~\cite{wang2024qwen2} and  LLaVa 1.5~\cite{liu2024llava}. Inspired by ADAPTVIS~\cite{chen2025spatial} and ViCrop~\cite{zhang2025mllms}, we additionally analyze the attention distribution over the image to gain deeper insights into the model's spatial reasoning capabilities. The results of Qwen2-VL are shown in \figurename~\ref{fig:case_study} with LLaVa 1.5~\cite{liu2024llava} results in Appendix~\ref{app:case}. 

The goal of the \textit{Path Reading} task is to evaluate whether the model can accurately attend to the corresponding regions in the image. 
Although both models fail to generate a completely correct answer, our findings reveal distinct attention patterns between the two models.
Qwen-2VL, compared to LLaVA-1.5, demonstrates a more accurate attention mechanism, as its correct responses consistently focus on the relevant regions of the image. 
This observation aligns with the findings of ADAPTVIS, which reported a strong correlation between successful spatial reasoning and the model’s ability to align its attention distribution with actual locations.
The difference between these models may be attributed to architectural enhancements in Qwen-2VL, such as the use of 2D-RoPE in the vision encoder and M-RoPE in the language model backbone.
These findings further support our analysis of position embedding in Sec~\ref{sec:model}.

The \textit{Planning} task requires more complex spatial reasoning capabilities.
Our results indicate that both Qwen-2VL and LLaVA-1.5 fail to generate correct path, while responses of Qwen-2VL are much more reasonable. 
The attention maps demonstrate that Qwen-2VL primarily focuses on object-level semantic information, such as `yellow star', rather than forming a higher-level understanding of the spatial layout.
To further explore these limitations, we introduced extra visual cues, as illustrated in the bottom section of \figurename~\ref{fig:case_study}.
Additionally, we employed the CoT prompting technique to investigate the reasoning steps. 
The results reveal that Qwen-2VL generates a reasonable path while ignoring the black boundary.
As shown in attention maps, Qwen-2VL also attends to the visual cues during its reasoning process.
However, it remains challenging when reasoning about the accurate spatial relationship, like `right to the red triangle'.

In summary, these observations indicate that improved recipes lead to better spatial reasoning capability.
Besides, current MLLMs predominantly focus on object-level information, aligning with most existing training data and paradigms. 
These findings support our argument for the need for new recipes tailored to enhance MLLMs’ spatial reasoning capabilities.

\vspace{-2mm}
\section{Conclusion}
\label{sec:conclusion}
\vspace{-2mm}
This position paper discusses a critical and timely gap of spatial reasoning capabilities in MLLMs.
By systematically dissecting the challenges in spatial reasoning, we underscore the urgent need for targeted research and development. 
Our analysis reveals that the components of current recipes-from training data to reasoning mechanisms-limit the spatial reasoning capability of MLLMs, not only constraining real-world applicability but also hindering the way to true intelligence.
Therefore, we call on the community to prioritize spatial reasoning and develop new recipes for MLLMs to effectively enhance their spatial reasoning capabilities.

\bibliography{reference}
\bibliographystyle{unsrtnat}

\newpage
\appendix

\section{Additional Results of Case study}
\label{app:case}

\begin{figure}
    \centering
    \includegraphics[width=0.8\linewidth]{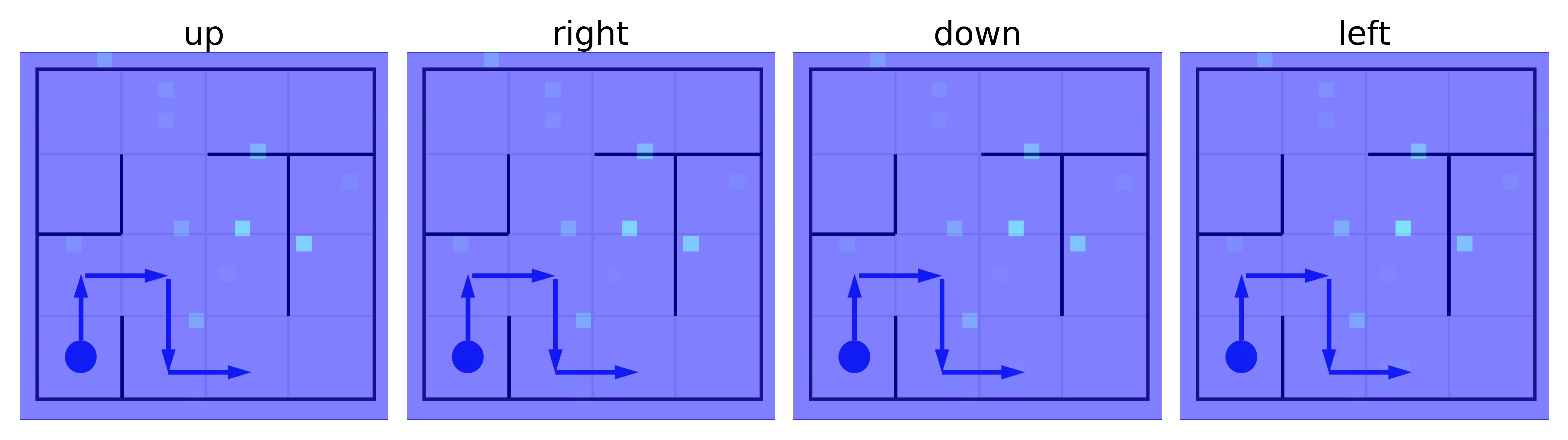}
    \vspace{-5mm}
    \caption{Attention maps of LLaVa 1.5 in the Path Reading task}
    \vspace{-2mm}
    \label{fig:app_llava}
\end{figure}

The attention maps of LLaVa 1.5 in the Path Reading task are shown in \figurename~\ref{fig:app_llava}.
The visualization illustrates that LLaVa 1.5 does not pay attention to the corresponding regions in the image when generating answers, which is completely different from Qwen-2VL.

The entire input and output of different tasks are provided in Table~\ref{apptab:qwen-prompt} and \ref{apptab:llava-prompt}.
In the case of Qwen-2VL, employing Chain-of-Thought reasoning in the absence of visual cues within planning task may lead to repetitive responses. without visual cues, the response will fall into repetition.
For LLaVa 1.5, regardless of the use of Chain of Thought (CoT) or visual cues, it tends to produce answers that are excessively complex and incorrect.
This issue may stem from its simpler recipe compared to Qwen-2VL.

\section{Experimental Setup in Case Study}
\label{app:setup}

We employ Qwen2-VL-7B-Instruct\footnote{\href{https://huggingface.co/Qwen/Qwen2-VL-7B-Instruct}{https://huggingface.co/Qwen/Qwen2-VL-7B-Instruct}} and LLaVa-1.5-7b\footnote{\href{https://huggingface.co/liuhaotian/llava-v1.5-7b}{https://huggingface.co/liuhaotian/llava-v1.5-7b}} in our case study. 
The input image will be resized to $448\times 448$ prior to being processed by the vision encoder. 
All experiments conducted in the case study utilize a single NVIDIA A100-80G GPU. We provide attention visualizations that are averaged across all heads and layers.
For each text token, the attention to all visual tokens is mapped back onto the original image for display.

\begin{table*}[]
\centering
\caption{The input and output of Qwen-2VL. }
\begin{tcolorbox}[title = {\textbf{Qwen-2VL}}]

\textbf{Task: Path Reading }

\textbf{Input:} 

This is a picture of a maze and there is a path colored in red. Each red arrow represents one action. There are 4 actions in total. The definitions of the actions are as below. 

Go up/left/down/right: move one grid space in the absolute up/left/down/right direction. 

The start point is the red circle. Provide the entire action sequence following the red path from the start point. 

\textbf{Response:} 

The action sequence is: go up, go left, go down, go right.
\\
\\
\textbf{Task: Planning }

\textbf{Input:} 

This is a picture of a maze. The definitions of the actions are as below. 

Go up/left/down/right: move one grid space in the absolute up/left/down/right direction. 

The start point is the red circle and the destination is the yellow star. Provide the action sequence from the start to the destination. 

\textbf{Response:} 

The agent moves up, then right, then up, then right, and finally up to reach the yellow star.
\\
\\
\textbf{Task: Planning with CoT}

\textbf{Input:}

This is a picture of a maze. The definitions of the actions are as below. 

Go up/left/down/right: move one grid space in the absolute up/left/down/right direction. 

The start point is the red circle and the destination is the yellow star. Provide the action sequence from the start to the destination. 

Let's think step by step.

\textbf{Response:} 

To solve this maze and reach the yellow star from the red circle, let's analyze the grid step by step:

1. Starting Point: The red circle is at the bottom left corner.

2 the yellow star is at the top right corner.

2 the yellow star is at the top right corner.

2 the yellow star is at the top right corner.

...
\\
\\
\textbf{Task: Planning with Visual Cues and CoT}

\textbf{Input:}

This is a picture of a maze. The definitions of the actions are as below. 

Go up/left/down/right: move one grid space in the absolute up/left/down/right direction. 

The start point is the red circle and the destination is the yellow star. Provide the action sequence from the start to the destination. 

Let's think step by step.

\textbf{Response:} 

1. Start at the red circle in the bottom left corner.

2. Move right one grid space up to the blue triangle.

3. Move right one grid space right to the red triangle.

4. Move one grid space up to the yellow star.

\end{tcolorbox}
\label{apptab:qwen-prompt}
\end{table*}

\begin{table*}[]
\centering
\caption{The input and output of LLaVa 1.5. }
\begin{tcolorbox}[title = {\textbf{LLaVa 1.5}}]

\textbf{Task: Path Reading }

\textbf{Input:} 

This is a picture of a maze and there is a path colored in red. Each red arrow represents one action. There are 4 actions in total. The definitions of the actions are as below. 

Go up/left/down/right: move one grid space in the absolute up/left/down/right direction. 

The start point is the red circle. Provide the entire action sequence following the red path from the start point. 

\textbf{Response:} 

1. Go up
2. Go right
3. Go down
4. Go left
\\
\\
\textbf{Task: Planning }

\textbf{Input:} 

This is a picture of a maze. The definitions of the actions are as below. 

Go up/left/down/right: move one grid space in the absolute up/left/down/right direction. 

The start point is the red circle and the destination is the yellow star. Provide the action sequence from the start to the destination. 

\textbf{Response:} 

Start at the red circle, go up, right, down, left, up, right, down, left, up, right, down, left, up, right, down, left, up, right, down, left, up, right, down, left, up, right, down, left, up, right, down, left, up, right, down, left, ...
\\
\\
\textbf{Task: Planning with CoT}

\textbf{Input:}

This is a picture of a maze. The definitions of the actions are as below. 

Go up/left/down/right: move one grid space in the absolute up/left/down/right direction. 

The start point is the red circle and the destination is the yellow star. Provide the action sequence from the start to the destination. 

Let's think step by step.

\textbf{Response:} 

To solve the maze and reach the yellow star, follow these steps:

1. Start at the red circle, which is the current position.\\
2. Go up one grid space.\\
3. Go left one grid space.\\
4. Go down one grid space.\\
5. Go right one grid space.\\
...\\
50. Go up one grid space
\\
\\
\textbf{Task: Planning with Visual Cues and CoT}

\textbf{Input:}

This is a picture of a maze. The definitions of the actions are as below. 

Go up/left/down/right: move one grid space in the absolute up/left/down/right direction. 

The start point is the red circle and the destination is the yellow star. Provide the action sequence from the start to the destination. 

Let's think step by step.

\textbf{Response:} 

To solve the maze, follow these steps:

1. Start at the red circle.\\
2. Go up one grid space.\\
3. Go left one grid space.\\
4. Go down one grid space.\\
...\\
51. Go left one grid space.

\end{tcolorbox}
\label{apptab:llava-prompt}
\end{table*}

\end{document}